\documentclass{sig-alternate-2013}

\usepackage{mathrsfs}
\usepackage{color}
\usepackage{graphicx}
\usepackage{caption}
\usepackage{subcaption}
\usepackage[normalem]{ulem}
\usepackage{algorithm}
\usepackage[noend]{algpseudocode}

\graphicspath{{./Fig/}}
\DeclareGraphicsExtensions{.pdf,.jpeg,.png,.jpg,.eps,.tif}

\DeclareMathOperator*{\argmin}{argmin}
\DeclareMathOperator*{\argmax}{argmax}

\usepackage[belowskip=-10pt,aboveskip=-0pt]{caption}
\usepackage[font=small,labelfont=bf]{caption}

\newfont{\mycrnotice}{ptmr8t at 7pt}
\newfont{\myconfname}{ptmri8t at 7pt}
\let\crnotice\mycrnotice%
\let\confname\myconfname%

\permission{Permission to make digital or hard copies of all or part of this work for personal or classroom use is granted without fee provided that copies are not made or distributed for profit or commercial advantage and that copies bear this notice and the full citation on the first page. Copyrights for components of this work owned by others than ACM must be honored. Abstracting with credit is permitted. To copy otherwise, or republish, to post on servers or to redistribute to lists, requires prior specific permission and/or a fee. Request permissions from Permissions@acm.org.}
\conferenceinfo{MM'15,}{October 26--30, 2015, Brisbane, Australia.} 
\copyrightetc{\copyright~2015 ACM. ISBN \the\acmcopyr}
\crdata{978-1-4503-3459-4/15/10\ ...\$15.00.\\
DOI: http://dx.doi.org/10.1145/2733373.2806334}

\begin{document}
\title{Coupled Support Vector Machines for Supervised \\Domain Adaptation}

\numberofauthors{1} 
\author{\alignauthor Hemanth Venkateswara$^*$, Prasanth Lade$^\dagger$,  Jieping Ye$^\ddagger$, Sethuraman Panchanathan$^*$ \\
\affaddr{$^*$Center for Cognitive Ubiquitous Computing, Arizona State Univeristy} \\ 
\affaddr{$^\dagger$ Bosch Research and Technology Center, Palo Alto} \affaddr{\hspace{5 mm}$^\ddagger$ University of Michigan, Ann Arbor} \\
\footnotesize
\email{hemanthv@asu.edu, prasanth.lade@us.bosch.com, jpye@umich.edu, panch@asu.edu}
}

\maketitle
\begin{abstract}
\small\baselineskip=7pt
Popular domain adaptation (DA) techniques learn a classifier for the target domain by sampling relevant data points from the source and combining it with the target data. We present a Support Vector Machine (SVM) based supervised DA technique, where the similarity between source and target domains is modeled as the similarity between their SVM decision boundaries. We couple the source and target SVMs and reduce the model to a standard single SVM. We test the Coupled-SVM on multiple datasets and compare our results with other popular SVM based DA approaches.
\end{abstract}




\section{Introduction and Motivation}
Supervised learning algorithms often make the implicit assumption that the test data is drawn from the same distribution as the training data. These algorithms become ineffective when such assumptions regarding the test data are violated. \textit{Transfer learning} techniques are applied to address these kinds of problems. Transfer learning involves extracting knowledge from one or more tasks or domains and utilizing (transferring) that knowledge to design a solution for a new task or domain \cite{bruzzone2010domain}. Domain adaptation (DA) is a special case of transfer learning where we handle data from different, yet correlated distributions. DA techniques transfer knowledge from the \textit{source} domain (distribution) to the \textit{target} domain (distribution), in the form of learned models and efficient feature representations, to learn effective classifiers on the target domain.
\begin{figure}
\centering
\includegraphics[width=0.49\textwidth]{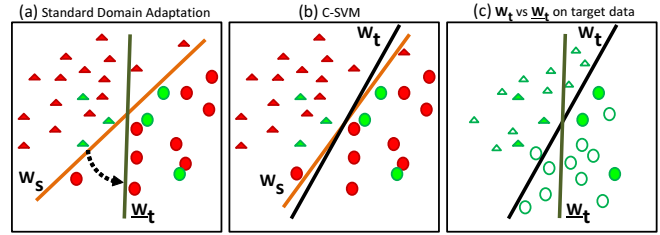}
\caption{(a) Standard SVM based DA. The source SVM $\mathbf{w}_s$ is perturbed to get $\uline{\mathbf{w}}_t$. (b) C-SVM, $\mathbf{w}_s$ and $\mathbf{w}_t$ are learned together. Training error could be high. (c) The C-SVM does not over fit. ${\color{red}Red}$ is source and ${\color{green}Green}$ is target data. Filled unfilled objects are train and test data respectively.}
\label{fig:c-svm}
\vspace{-1.2em}
\end{figure}
In this work we consider the problem of supervised DA where we use labeled samples from the source domain along with a limited number of labeled samples from the target domain, to learn a classifier for the target domain. We propose a Coupled linear Support Vector Machine (C-SVM) model that simultaneously estimates linear SVM decision boundaries $\mathbf{w}_s$ and $\mathbf{w}_t$, for the source and target training data respectively. Using a technique termed as \textit{instance matching}, researchers sample source data points such that the difference between the means of the sampled source and target data is minimized \cite{duan2012domain}, \cite{long2014transfer}. Our intuition behind the C-SVM is along similar lines, where we penalize the difference between $\mathbf{w}_s$ and $\mathbf{w}_t$. Since the SVM decision boundaries are a linear combination of the data points, penalizing the difference between $\mathbf{w}_s$ and $\mathbf{w}_t$, can be viewed as penalizing the difference between the weighted means of the source and target data points.

Figure(\ref{fig:c-svm}a), illustrates standard SVM based DA where $\mathbf{w}_s$ is first learned on the source and is subsequently perturbed to obtain the target ($\mathbf{\underline{w}}_t$). The perturbed SVM $\mathbf{\underline{w}}_t$ could be very different from $\mathbf{w}_s$ and can overfit the target training data. Figure(\ref{fig:c-svm}b), depicts the C-SVM, where $\mathbf{w}_s$ and $\mathbf{w}_t$, are learned simultaneously. The source SVM $\mathbf{w}_s$, provides an anchor for the target SVM $\mathbf{w}_t$. The difference between $\mathbf{w}_s$ and $\mathbf{w}_t$ is modeled based on the difference between the source and target domains. In addition, the C-SVM trades training error for generalization as illustrated in Figure(\ref{fig:c-svm}c). In this paper, we formulate a coupled SVM problem to estimate $\mathbf{w}_s$ and $\mathbf{w}_t$ and reduce it to a single SVM problem that can be solved with standard quadratic optimization. We test our model and report recognition accuracies on various datasets of objects, hand-written digits, facial expressions and activities.

\section{Related Work and Our Method}
\label{label:relatedWork}
In this section we will discuss some of the SVM based DA techniques closely related to the C-SVM. Support Vector Machines have been extensively used for DA in the past. Daum{\'e} \cite{daumeiii2007}, modeled augmented features with a heuristic kernel. Bruzzone and Marconcini \cite{bruzzone2010domain}, proposed an unsupervised method (DASVM) to adapt a SVM learned on the source domain to the unlabeled data in the target domain in an iterative manner. Adapt-SVM is another technique closely related to our method, where Yang et al. \cite{yang2007adapting} and Li \cite{li2007regularized}, learn a SVM on the target by minimizing the classification error on the target data while also reducing the discrepancy between the source and target SVMs. We differ from this method by learning the source and target SVMs simultaneously. Aytar and Zisserman \cite{aytar2011tabula}, extend this framework to the Projective Model Transfer SVM that relaxes the transfer induced by the Adapt-SVM. Hoffman et al. (MMDT) \cite{hoffman2013}, learn a single SVM model for the source and the transformed target data. The target data is transformed by a transformation matrix that is learned in an optimization framework along with the SVM. Duan et al. (AMKL) \cite{duan2012domain} implement a multiple kernel method where multiple base kernel classifiers are combined with a pre-learned average classifier obtained from fusing multiple nonlinear SVMs. Unlike in C-SVM where the similarity between source and target is learned by the model, Widmer et al. \cite{widmer2012efficient} use a similar approach to solve multitask problems using graph Laplacians to model task similarity. We believe the C-SVM holds a unique position in this wide array of SVM solutions for DA. The C-SVM trains a linear SVM for both the source and target domains simultaneously, thereby minimizing the chances of over-fitting, especially when there are very few labeled samples from the target domain.

\section{Problem Specification}  
We outline the problem as follows. Let $\{(\mathbf{x}_i^s, y_i^s)\}_{i=1}^{N_s} \subset \mathscr{S}$, where $\mathscr{S}$ is the \textit{source domain}, $\mathbf{x}_i^s \in \mathbb{R}^d$, are data points and $y_i^s \in \{-1, +1\}$, are their labels. Along similar lines, $\{(\mathbf{x}_i^t, y_i^t)\}_{i=1}^{N_t} \subset \mathscr{T}$, where $\mathscr{T}$ is the \textit{target domain}. 
\subsection{Coupled-SVM Model}
The goal is to learn a target classifier $\mathit{f_t}$, that generalizes to a larger subset of $\mathscr{T}$ and does not over fit the target training data $\{(\mathbf{x}_i^t, y_i^t)\}_{i=1}^{N_t}$. The catch here is that the number of labeled target data points is small and $N_t \ll N_s$. We therefore include the source data and learn the source classifier $\mathit{f_s}(\mathbf{x}) = \mathbf{w}_s^\top\mathbf{x} + b_s$ to provide an anchor point for $\mathit{f_t} =  \mathbf{w}_t^\top\mathbf{x} + b_t$. The source and target SVM decision boundaries are $\{\mathbf{w}_s, b_s\}$ and $\{\mathbf{w}_t, b_t\}$ respectively. To simplify notation we re-define, $\mathbf{w}_s \leftarrow [\mathbf{w}_s^\top, b_s]^\top$ and $\mathbf{w}_t \leftarrow [\mathbf{w}_t^\top, b_t]^\top$ and account for the bias by re-defining, $\mathbf{x}_i^s \leftarrow [\mathbf{x}_i^{s^\top}, 1]^\top$ and $\mathbf{x}_i^t \leftarrow [\mathbf{x}_i^{t^\top}, 1]^\top$. Incorporating these definitions, the Coupled-SVM can be detailed as follows,
\begin{flalign}
\{\mathbf{w}_s^*,\mathbf{w}_t^*\}=&~\argmin\limits_{\mathbf{w}_s,\mathbf{w}_t}~\frac{1}{2}\lambda||\mathbf{w}_s-\mathbf{w}_t||_2^2 + 
\frac{1}{2}||\mathbf{w}_s||_2^2 + \frac{1}{2}||\mathbf{w}_t||_2^2 \notag\\
&+~C_s\sum_i^{N_s}\xi_i^s + C_t\sum_i^{N_t}\xi_i^t \notag\\
\text{s.t.},~~y_i^s(\mathbf{w}_s^\top &\mathbf{x}_i^s) \geq 1-\xi_i^s, ~\xi_i^s\geq 0, ~~ i\in [1,\ldots,N_s]\notag\\
y_i^t(\mathbf{w}_t^\top &\mathbf{x}_i^t) \geq 1-\xi_i^t, ~\xi_i^t\geq 0, ~~ i\in [1,\ldots,N_t]
\label{modelEq}
\end{flalign}
Equation (\ref{modelEq}) is a variation of a standard linear SVM with two decision boundaries and an additional term relating the two boundaries. The first term captures the similarity(dissimilarity) between the source and target domains as the difference between the decision boundaries. $\lambda$ controls the importance of this difference. The 2nd and 3rd term are the SVM regularizers. The 4th and 5th terms capture training loss, where $C_s$ and $C_t$ control the importance of the source and target misclassification respectively. 
\subsection{Solution}
To simplify notation, we define a new set of variables based on the earlier ones. We concatenate the two SVM boundaries into a single variable, $\mathbf{w} \in \mathbb{R}^{2(d+1)}$ defined as, $\mathbf{w} \leftarrow [\mathbf{w}_s^\top, \mathbf{w}_t^\top]^\top$. 
The individual SVMs $\mathbf{w}_s$ and $\mathbf{w}_t$ can be extracted from $\mathbf{w}$ using permutation matrices $I_s \in \mathbb{R}^{(d+1)\times 2(d+1)}$ and $I_t \in \mathbb{R}^{(d+1)\times 2(d+1)}$, where $I_s$ and $I_t$ are binary matrices such that, $I_s\mathbf{w} = \mathbf{w}_s$ and $I_t\mathbf{w} = \mathbf{w}_t$. 
For example, let $\mathbf{v}_1 = [a,b]^\top$, $\mathbf{v}_2 = [c,d]^\top$ and $\mathbf{v} \leftarrow [\mathbf{v}_1^\top, \mathbf{v}_2^\top]^\top $. Then the permutation matrices $I_{v_1}$ and $I_{v_2}$ such that, $I_{v_1}\mathbf{v} = \mathbf{v}_1$ and $I_{v_2}\mathbf{v} = \mathbf{v}_2$, are given by, 
$I_{v_1} = \left[ \begin{smallmatrix} 1 & 0 & 0 & 0\\ 0 & 1 & 0 & 0 \end{smallmatrix} \right]$, and 
$I_{v_2} = \left[ \begin{smallmatrix} 0 & 0 & 1 & 0 \\ 0 & 0 & 0 & 1 \end{smallmatrix} \right]$.
We also define new variables $(\mathbf{x}_i, y_i, c_i), i\in\{1,\ldots,(N_s+N_t)\}$, where, $\mathbf{x}_i \in \mathbb{R}^{2(d+1)}$ and $y_i \in \{-1, +1\}$ are the new data points and $c_i \in \{C_s, C_t\}$,
\begin{flalign}
\mathbf{x}_i \leftarrow
\begin{cases}
~[\mathbf{x}_i^{s^\top}, \mathbf{0}^\top]^\top \quad\quad\quad\quad~~~ 1\leq i \leq N_s\\
[\mathbf{0}^\top, \mathbf{x}_{i - N_s}^{t^\top}]^\top \quad (N_s+1)\leq i \leq (N_s+N_t)
\end{cases}
\label{label:Eqx}
\end{flalign}
where, $\mathbf{0} \in \mathbb{R}^{d+1}$ is a vector of zeros. Similarly, 
\begin{flalign}
\{y_i, c_i\} \leftarrow
\begin{cases}
~~\{y_i^s, C_s\} \quad\quad\quad\quad\quad 1\leq i \leq N_s\\
\{y_{i - N_s}^t, C_t\} \quad (N_s+1)\leq i \leq (N_s+N_t)
\end{cases}
\label{label:EqyC}
\end{flalign}
For ease of derivation, we consider the linearly separable SVM and get rid of $c_i$ (we will re-introduce it later). The minimization problem in Equation(\ref{modelEq}) can now be re-formulated as,
\begin{flalign}
\{&\mathbf{w}^*\}~=~\argmin\limits_{\mathbf{w}}~\frac{1}{2}\lambda||I_{st}\mathbf{w}||_2^2+\frac{1}{2}||\mathbf{w}||_2^2\notag\\
&\text{s.t.}, \quad y_i(\mathbf{w}^\top\mathbf{x}_i)\geq1\quad i\in\{1,\ldots,(N_s+N_t)\}
\label{modelEq2}
\end{flalign}
where we have defined, $I_{st} \leftarrow (I_s-I_t)$ and used $I_s\mathbf{w} = \mathbf{w}_s$, $I_t\mathbf{w} = \mathbf{w}_t$ for the first term. For the second term, we have used $\frac{1}{2}||\mathbf{w}_s||_2^2 + \frac{1}{2}||\mathbf{w}_t||_2^2 = \frac{1}{2}||\mathbf{w}||_2^2$. We introduce Lagrangian variables $\{\alpha_i\}$ to solve the problem,
\begin{flalign}
L(\mathbf{w}, \boldsymbol{\alpha})=&\frac{1}{2}\lambda||I_{st}\mathbf{w}||_2^2+\frac{1}{2}||\mathbf{w}||_2^2 -\sum_i^{N_s+N_t}\alpha_i\big(y_i(\mathbf{w}^\top\mathbf{x}_i-1)\big)
\label{modelLagrangian}
\end{flalign}
We need to minimize the Lagrangian w.r.t $\mathbf{w}$ and maximize w.r.t to $\boldsymbol{\alpha}$. We optimize first w.r.t $\mathbf{w}$ by setting the derivative $\frac{\partial L(\mathbf{w}, \boldsymbol{\alpha})}{\partial\mathbf{w}} = 0 $ and get, $\mathbf{w} = D\sum_i^{N_s+N_t}\alpha_i y_i\mathbf{x}_i$ where, $I$ is an identity matrix and, $D = (I+\lambda I_{st}^\top I_{st})^{-1}$. 
By the nature of our permutation matrices $I_s$ and $I_t$,  $(I+\lambda I_{st}^\top I_{st})$ is full rank and therefore, $D$ exists. We define $\mathbf{v} \leftarrow\sum_i^{N_s+N_t}\alpha_i y_i\mathbf{x}_i$ and substitute for $\mathbf{w}$ in Equation(\ref{modelLagrangian}) to arrive at the SVM dual form which we need to maximize,
\begin{flalign}
\{\boldsymbol{\alpha}^{*}\} &=\argmax\limits_{\boldsymbol{\alpha}}
\frac{1}{2}\mathbf{v}^\top D^\top I_{st}^\top I_{st} D\mathbf{v} + \frac{1}{2}\mathbf{v}^\top D^\top D\mathbf{v}
-\mathbf{v}^\top D\mathbf{v} + \mathbf{1}^\top\boldsymbol{\alpha}\notag\\
&=\argmax\limits_{\boldsymbol{\alpha}}
~\mathbf{v}^\top(\frac{1}{2} D^\top(I + I_{st}^\top I_{st})D - D)\mathbf{v} + \mathbf{1}^\top\alpha\notag\\
&=\argmin\limits_{\boldsymbol{\alpha}}
~\frac{1}{2}\mathbf{v}^\top D\mathbf{v} - \mathbf{1}^\top\alpha\notag ~~\text{($D$ is symmetric)}\\
&=\argmin\limits_{\boldsymbol{\alpha}}
~\frac{1}{2}\mathbf{\alpha}^\top\mathcal{Q}\mathbf{\alpha} - \mathbf{1}^\top\alpha
\label{svmDualEq}
\end{flalign}

\noindent Equation(\ref{svmDualEq}) is the standard SVM dual where $\mathcal{Q}_{ij} = y_iy_j\mathbf{x}_i^\top D\mathbf{x}_j$ and $\mathbf{1}$ is a vector of $1s$. To use any of the standard SVM libraries, we can set $\tilde{\mathbf{x}}_i = D^{1/2}\mathbf{x}_i$. Then $\mathcal{Q}_{i,j} = y_iy_j\tilde{\mathbf{x}}_i^\top\tilde{\mathbf{x}}_j$. The decision boundary in the space of $\tilde{\mathbf{x}}_i$, is given by, $\tilde{\mathbf{w}} = \sum_i\alpha_iy_i\tilde{\mathbf{x}}_i$. The decision boundary in the space of $\mathbf{x}_i$ is given by $\mathbf{w} = D\sum_i\alpha_iy_i\mathbf{x}_i$. Therefore $\mathbf{w} = D^{1/2}\tilde{\mathbf{w}}$. We re-introduce the slack variables as constraints $0\leq\alpha_i\leq c_i$. We can easily extend the algorithm to the multi-class setting using one-vs-one or one-vs-all settings. Once $\mathbf{w}$ is estimated, $I_s\mathbf{w} = \mathbf{w}_s$ and $I_t\mathbf{w} = \mathbf{w}_t$ is used to get the source and target SVMs.

\section{Experiments}
In this section we discuss the extensive experiments we conducted to study the C-SVM model. We first outline the different datasets and their domains. We then outline the DA algorithms we compare against. Finally, we report the experimental details and our results.

\subsection{Data Preparation}
For our experiments, we consider multiple datasets from different applications and also test the C-SVM with different kinds of features. For all the experiments (except \textit{Office-Caltech}) we use the following setting. For the training data, we sample $20$ examples from the source domain and $3$ examples from the target domain from every category. The test data is the remaining examples in the target domain not used for training.\\
\noindent\textbf{\textit{Office-Caltech} datasets}: For this experiment we borrow the dataset and the experimental setup outlined in \cite{gong2012geodesic}. The \textit{Office} dataset consists of three domains, \texttt{Amazon}, \texttt{Dslr} and \texttt{Webcam}. The \textit{Caltech256} dataset has one domain, \texttt{Caltech}. All the domains consist of a set of common categories viz., \textit{\{back-pack, bike, calculator, headphones, computer-keyboard, laptop, monitor, computer-mouse, coffee-mug, video-projector\}}. We use the $800$ dimension SURF-BoW features that are provided by\cite{gong2012geodesic} for our experiments. We follow the experimental setup outlined in \cite{gong2012geodesic}. For the training data, we sample $8$ examples from the source domain (for \texttt{Amazon} we use $20$) and $3$ examples from the target domain.\\
\noindent\textbf{\textit{MNIST-USPS} datasets}: The \textit{MNIST} and \textit{USPS} datasets are benchmark datasets for handwritten digit recognition. These datasets contain gray scale images of digits from $0$ to $9$. For our experiments, we have considered a subset of these datasets ($2000$ images from \textit{MINST} and $1800$ images from \textit{USPS}) based on \cite{long2014transfer}. We refer to these domains as \texttt{MNIST} and \texttt{USPS} respectively. The images are resized to $16\times 16$ pixels and represented as vectors of length 256.\\
\noindent\textbf{\textit{CKPlus-MMI} dataset}: The \textit{CKPlus}\cite{ckplusDataset} and \textit{MMI}\cite{pantic2005web} are benchmark facial expression recognition datasets. We select 6 categories viz., \{\textit{anger}, \textit{disgust}, \textit{fear}, \textit{anger}, \textit{happy}, \textit{sad}, and \textit{surprise}\}, from frames with the most intense expression (peak frames) from every facial expression video sequence to get around 1500 images for each dataset with around 250 images per category. We refer to these domains as \texttt{CKPlus} and \texttt{MMI}. We extract deep convolutional neural network based generic features which have shown astounding results across multiple applications \cite{razavian2014cnn}. We therefore decided to use an `off-the-shelf' feature extractor developed by Simonyan and Zisserman \cite{simonyan2014very}. We used the output of the first fully connected layer from the 16 weight layer model as features with dimension 4096 which were then reduced to 100 using PCA.\\
\noindent\textbf{\textit{HMDB51-UCF50} dataset}: We pooled $11$ common categories of activity from \textit{HMDB51}\cite{Kuehne11} and \textit{UCF50}\cite{reddy2013recognizing}. The categories from \textit{UCF50} are, \{\textit{BaseballPitch(throw)}, \textit{Basketball(shoot\_ball)}, \textit{Biking(ride\_bike)}, \textit{Diving(dive)}, \textit{Fencing}\\\textit{(fencing)}, \textit{GolfSwing(golf)}, \textit{HorseRiding(ride\_horse)}, \textit{PullUps} \textit{(pullup)}, \textit{PushUps(pushup)}, \textit{Punch(punch)}, \textit{WalkingWithDog\\(walk)}\}. The category names from \textit{HMDB51} are in parenthesis. We refer to these domains as \texttt{HMDB51} and \texttt{UCF50}. We extract state-of-the-art \textbf{HOG}, \textbf{HOF}, \textbf{MBHx} and \textbf{MBHy} descriptors from the videos according to \cite{Kantorov}. We pool the descriptors into one grid $\textbf{1x1x1}$, and estimate Fisher Vectors with $K=256$ Gaussians. The dimension of these Fishers Vectors is $202,752$. We apply PCA and reduce the dimension to $100$.
\begin{table*}[t]
\centering
\tiny
\resizebox{6.8in}{!}{%
\begin{tabular}{|c|c|c|c|c|c|c|}
				\hline
  \textbf{Expt.}   & \textbf{SVM(T)} & \textbf{SVM(S)}& \textbf{SVM(S+T)}&  \textbf{MMDT} 	 & \textbf{AMKL} & \textbf{C-SVM}  \\ \hline\hline
A$\rightarrow$ W(1) & 56.06$\pm$0.95 & 37.36$\pm$1.19 & 51.26$\pm$1.19 & 64.87$\pm$1.26 & {\color{red}67.85$\pm$1.06} & 66.40$\pm$1.09\\\hline
A$\rightarrow$ D(2) & 43.15$\pm$0.78 & 37.64$\pm$0.96 & 47.56$\pm$0.99 & 54.41$\pm$1.00 & 56.22$\pm$0.89 & {\color{red}57.13$\pm$0.98}\\\hline
W$\rightarrow$ A(3) & 44.39$\pm$1.18 & 32.03$\pm$0.90 & 44.87$\pm$0.59 & 50.54$\pm$0.82 & 52.96$\pm$0.57 & {\color{red}53.97$\pm$0.42}\\\hline
W$\rightarrow$ D(4) & 45.20$\pm$1.34 & 61.06$\pm$0.86 & 65.39$\pm$0.89 & 62.48$\pm$0.98 & {\color{red}75.95$\pm$0.94} & 68.27$\pm$0.86\\\hline
D$\rightarrow$ A(5) & 42.17$\pm$1.03 & 31.48$\pm$0.65 & 46.17$\pm$0.44 & 50.45$\pm$0.75 & 52.36$\pm$0.57 & {\color{red}54.10$\pm$0.55}\\\hline
D$\rightarrow$ W(6) & 54.91$\pm$0.80 & 69.81$\pm$1.06 & 76.19$\pm$0.64 & 74.34$\pm$0.66 & {\color{red}85.94$\pm$0.44} & 77.17$\pm$0.46\\\hline
A$\rightarrow$ C(7) & 26.62$\pm$0.60 & 38.61$\pm$0.50 & 42.46$\pm$0.39 & 39.67$\pm$0.50 & {\color{red}44.92$\pm$0.46} & {\color{red}44.74$\pm$0.57}\\\hline
W$\rightarrow$ C(8) & 25.82$\pm$0.78 & 26.67$\pm$0.59 & 34.53$\pm$0.76 & 34.86$\pm$0.79 & {\color{red}39.20$\pm$0.57} & {\color{red}39.77$\pm$0.59}\\\hline
D$\rightarrow$ C(9) & 26.88$\pm$0.74 & 25.74$\pm$0.47 & 34.68$\pm$0.67 & 35.82$\pm$0.75 & {\color{red}41.12$\pm$0.44} & {\color{red}41.27$\pm$0.51}\\\hline
C$\rightarrow$ A(10) & 43.52$\pm$1.07 & 36.22$\pm$0.82 & 47.75$\pm$0.60 & 51.10$\pm$0.76 & {\color{red}55.98$\pm$0.58} & {\color{red}55.56$\pm$0.76}\\\hline
C$\rightarrow$ W(11) & 55.49$\pm$1.02 & 29.72$\pm$1.54 & 51.28$\pm$1.23 & 62.94$\pm$1.11 & {\color{red}68.70$\pm$1.07} & 67.74$\pm$1.05\\\hline
C$\rightarrow$ D(12) & 43.07$\pm$1.47 & 32.56$\pm$1.03 & 47.68$\pm$1.17 & 52.56$\pm$0.97 & 58.82$\pm$0.83 & {\color{red}59.72$\pm$1.01}\\\hline
M$\rightarrow$ U(13) & 70.73$\pm$0.41 & 38.89$\pm$0.61 & 64.36$\pm$0.41 & 68.96$\pm$0.43 & {\color{red}79.56$\pm$0.30} & 76.02$\pm$0.34\\\hline
U$\rightarrow$ M(14) & 58.23$\pm$0.39 & 21.67$\pm$0.33 & 38.43$\pm$0.36 & 48.29$\pm$0.32 & {\color{red}63.80$\pm$0.32} & {\color{red}63.25$\pm$0.31}\\\hline
K$\rightarrow$ I(15) & {\color{red}33.31$\pm$0.27} & 13.30$\pm$0.15 & 25.83$\pm$0.31 & 18.28$\pm$0.42 & 31.87$\pm$0.29 & {\color{red}33.10$\pm$0.29}\\\hline
I$\rightarrow$ K(16) & 45.65$\pm$0.47 & 19.47$\pm$0.55 & 25.63$\pm$0.35 & 21.33$\pm$0.81 & 43.59$\pm$0.50 & {\color{red}48.54$\pm$0.47}\\\hline
H$\rightarrow$ F(17) & 28.94$\pm$0.26 & 17.45$\pm$0.17 & 23.00$\pm$0.19 & 29.05$\pm$0.23 & 33.06$\pm$0.23 & {\color{red}35.89$\pm$0.25}\\\hline
F$\rightarrow$ H(18) & 18.64$\pm$0.19 & 16.99$\pm$0.16 & 19.58$\pm$0.17 & 22.28$\pm$0.18 & {\color{red}24.28$\pm$0.16} & {\color{red}24.41$\pm$0.19}\\\hline
\end{tabular}
}
\caption[Table caption text]{Recognition accuracies (\%) on the object, digit, facial expression abd activity datasets across multiple algorithms. \{\texttt{Amazon}(A), \texttt{Webcam}(W), \texttt{Dslr}(D), \texttt{Caltech}(C), \texttt{MNIST}(M), \texttt{USPS}(U), \texttt{CKPlus}(K), \texttt{MMI}(I), \texttt{HMDB51}(H), \texttt{UCF50}(F)\}. A$\rightarrow$W implies A is source domain and W is target domain. The best results are highlighted in red.}
\label{table:barGraph}
\end{table*}
\subsection{Existing Methods}
We compare our method with existing supervised DA techniques based on SVMs. 
\textbf{SVM(T)} (Linear SVM with training data from target domain), 
\textbf{SVM(S)} (Linear SVM with training data from source domain), 
\textbf{SVM(S+T)} (Linear SVM with union of source and target domain training data), 
\textbf{MMDT} (The Max-Margin Domain Transform \cite{hoffman2013}), 
\textbf{AMKL} (The Adaptive Multiple Kernel Learning \cite{duan2012domain}), and 
\textbf{C-SVM} (Coupled SVM algorithm).
\subsection{Experimental Details and Results}
We conducted $18$ experiments with different combinations of datasets. Table(\ref{table:barGraph}) depicts the results comparing multiple algorithms. For the \textit{Office-Caltech} dataset, the results are averaged across $20$ splits of data and $100$ splits for the rest of the experiments. The results for SVM(S) demonstrate the fact that although the datasets consist of the same categories, the domains have different distributions of data points. This is also highlighted by the success of SVM(T) even with few labeled training data points. The naive union of the source and target training data is in some cases beneficial but not always, as illustrated by SVM(S+T). Amongst the algorithms we have compared with, AMKL is on par with C-SVM in terms of performance. There is little to choose in terms of performance accuracies between the two. However, C-SVM is the easier and simpler solution as it is a standard linear SVM unlike AMKL, which is a multiple kernel based method.
\begin{figure*}[t]
        \centering
        \begin{subfigure}[b]{0.3\textwidth}
                \includegraphics[trim = 5mm 0mm 3mm 6mm, clip, width=2.3in, height=1.5in]{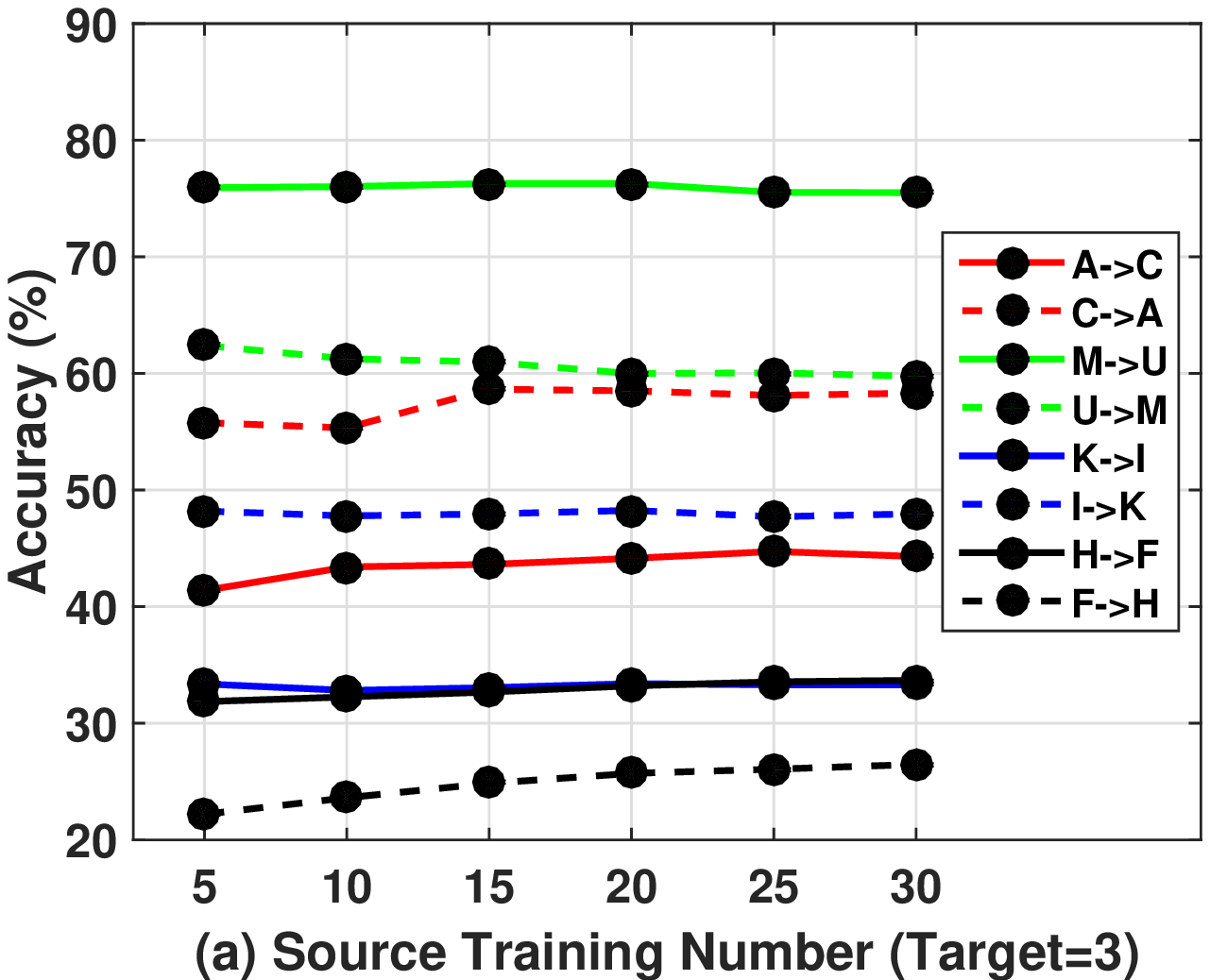}
                \label{fig:Source}
        \end{subfigure}%
        \quad 
        \begin{subfigure}[b]{0.3\textwidth}
                \includegraphics[trim = 5mm 0mm 3mm 6mm, clip, width=2.3in, height=1.5in]{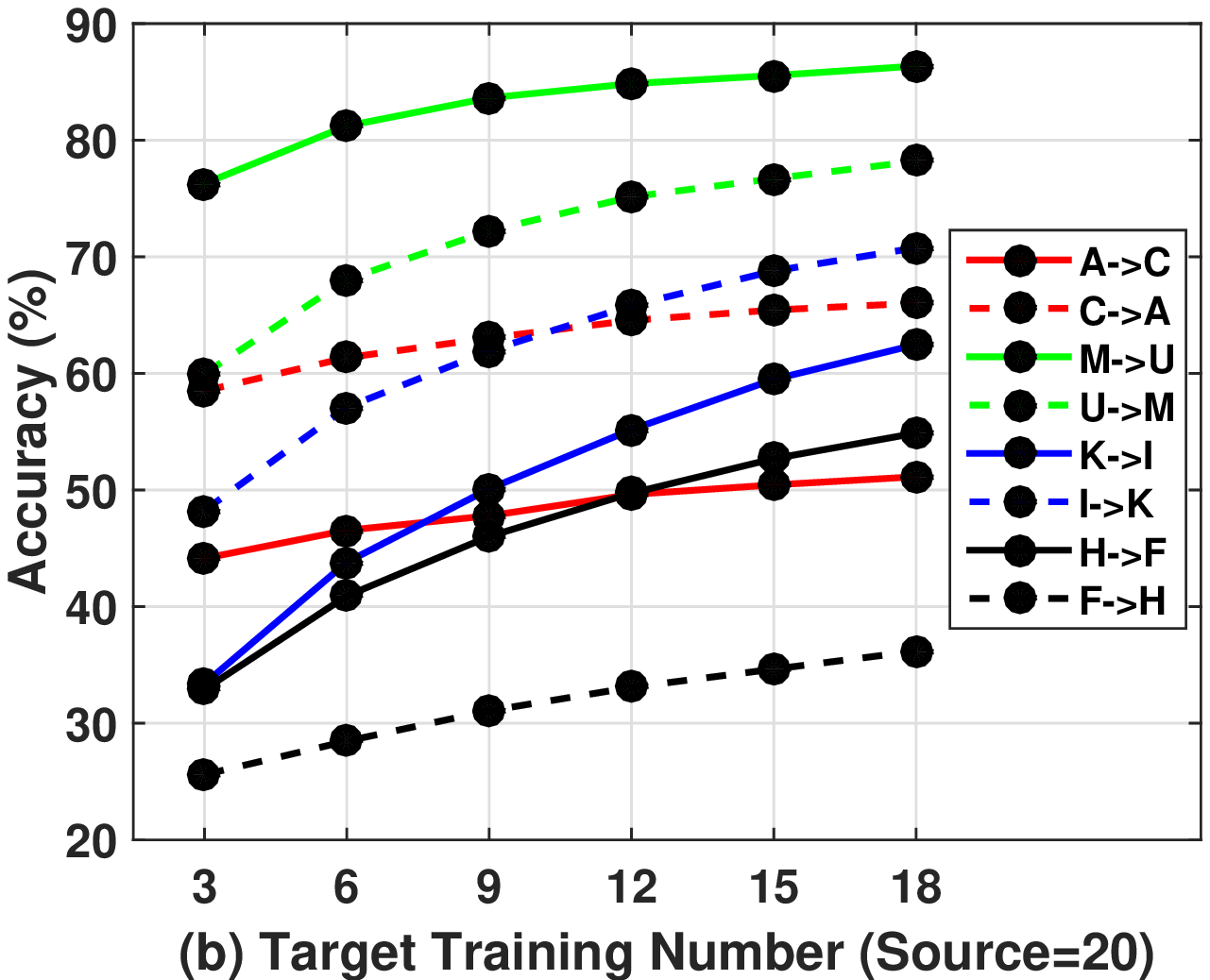}
                \label{fig:Target}
        \end{subfigure}
        \quad 
        \begin{subfigure}[b]{0.3\textwidth}
                \includegraphics[trim = 5mm 0mm 3mm 6mm, clip, width=2.3in, height=1.5in]{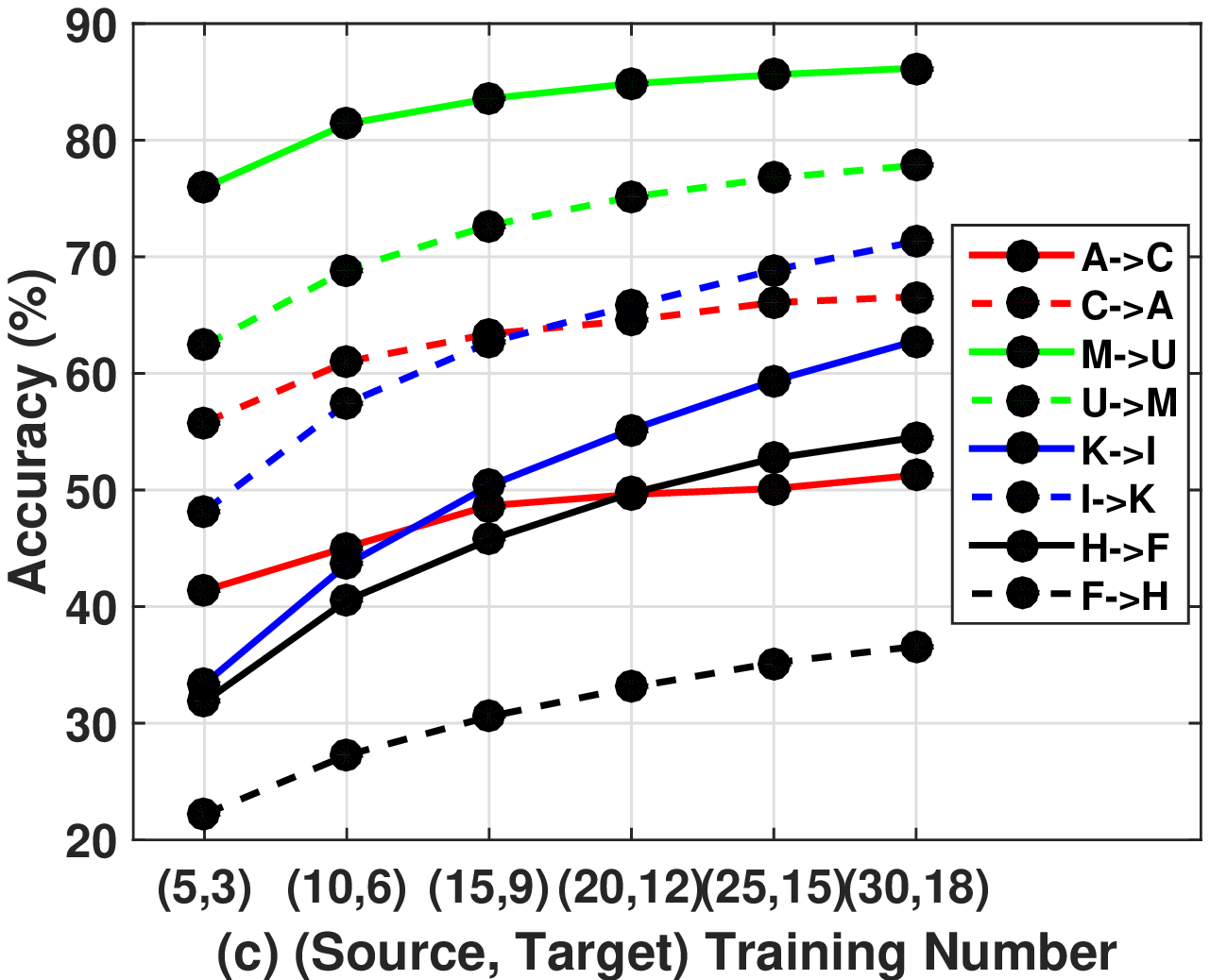}
                \label{fig:SourceTarget}
        \end{subfigure}
        \caption{Average recognition accuracy (\%) across different experiments varying number of training examples in source and target.}
\label{fig:experiments}
\vspace{-0.5em}
\end{figure*}
In all of these experiments we apply leave-one-out cross validation across the training target data to determine the best values of the parameters $\{C_s, C_t, \lambda\}$. 
We also studied the C-SVM by varying the number of samples available for training. We dropped the \texttt{Webcam} and \texttt{Dslr} datasets as they have fewer number of data points. Figure(\ref{fig:experiments}a) illustrates that increasing the number of source training data points, does not affect the test accuracies. The SVM relies on support vectors to estimate the source decision boundary, and additional source training data does not modify the source boundaries by much. The effect of additional target training data is comparatively more pronounced in Figure(\ref{fig:experiments}b) which is intuitive. By far, the most interesting is Figure(\ref{fig:experiments}c). Increasing both source and target training data numbers is nearly comparable to increasing only the number of target training data points. Source training data does not contribute to the target SVM after a threshold number of training data points.
\section{Conclusions} 
The C-SVM is elegant, efficient and easy to implement. We plan to extend this work to study nonlinear adaptations in the future. We would like to model classifier similarity in an infinite dimensional (kernel) space and also contemplate on the idea of unsupervised DA.
\vspace*{-1mm}
\scriptsize
\section{Acknowledgments}
This material is based upon work supported by the National Science Foundation (NSF) under Grant No:1116360. Any opinions, findings, and conclusions or recommendations expressed in this material are those of the authors and do not necessarily reflect the views of the NSF.
\vspace*{-1mm}
\scriptsize

\end{document}